# A swarm optimization algorithm inspired in the behavior of the social-spider

Erik Cuevas[1], Miguel Cienfuegos, Daniel Zaldívar, Marco Pérez-Cisneros

Departamento de Electrónica
Universidad de Guadalajara, CUCEI
Av. Revolución 1500, Guadalajara, Jal, México
{[1]erik.cuevas, daniel.zaldivar, marco.perez}@cucei.udg.mx

**Abstract**

Swarm intelligence is a research field that models the collective behavior in swarms of insects or animals. Several algorithms arising from such models have been proposed to solve a wide range of complex optimization problems. In this paper, a novel swarm algorithm called the Social Spider Optimization (SSO) is proposed for solving optimization tasks. The SSO algorithm is based on the simulation of cooperative behavior of social-spiders. In the proposed algorithm, individuals emulate a group of spiders which interact to each other based on the biological laws of the cooperative colony. The algorithm considers two different search agents (spiders): males and females. Depending on gender, each individual is conducted by a set of different evolutionary operators which mimic different cooperative behaviors that are typically found in the colony. In order to illustrate the proficiency and robustness of the proposed approach, it is compared to other well-known evolutionary methods. The comparison examines several standard benchmark functions that are commonly considered within the literature of evolutionary algorithms. The outcome shows a high performance of the proposed method for searching a global optimum with several benchmark functions.

## 1. Introduction

The collective intelligent behavior of insect or animal groups in nature such as flocks of birds, colonies of ants, schools of fish, swarms of bees and termites have attracted the attention of researchers. The aggregative conduct of insects or animals is known as swarm behavior. Entomologists have studied this collective phenomenon to model biological swarms while engineers have applied these models as a framework for solving complex real-world problems. This branch of artificial intelligence which deals with the collective behavior of swarms through complex interaction of individuals with no supervision is frequently addressed as swarm intelligence. Bonabeau defined swarm intelligence as ''any attempt to design algorithms or distributed problem solving devices inspired by the collective behavior of the social insect colonies and other animal societies" [1]. Swarm intelligence has some advantages such as scalability, fault tolerance, adaptation, speed, modularity, autonomy and parallelism [2].

The key components of swarm intelligence are self-organization and labor division. In a self-organizing system, each of the covered units responds to local stimuli individually and may act together to accomplish a global task, via a labor separation which avoids a centralized supervision. The entire system can thus efficiently adapt to internal and external changes.

Several swarm algorithms have been developed by a combination of deterministic rules and randomness, mimicking the behavior of insect or animal groups in nature. Such methods include the social behavior of bird flocking and fish schooling such as the Particle Swarm Optimization (PSO) algorithm [3], the cooperative behavior of bee colonies such as the Artificial Bee Colony (ABC) technique [4], the social foraging behavior of bacteria such as the Bacterial Foraging Optimization Algorithm (BFOA) [5], the simulation of the herding behavior of krill individuals such as the Krill Herd (KH) method [6], the mating behavior of firefly insects such as the Firefly (FF) method [7] and the emulation of the lifestyle of cuckoo birds such as the Cuckoo Optimization Algorithm (COA) [8].

---

[1] Corresponding author, Tel +52 33 1378 5900, ext. 27714, E-mail: erik.cuevas@cucei.udg.mx

In particular, insect colonies and animal groups provide a rich set of metaphors for designing swarm optimization algorithms. Such cooperative entities are complex systems that are composed by individuals with different cooperative-tasks where each member tends to reproduce specialized behaviors depending on its gender [9]. However, most of swarm algorithms model individuals as unisex entities that perform virtually the same behavior. Under such circumstances, algorithms waste the possibility of adding new and selective operators as a result of considering individuals with different characteristics such as sex, task-responsibility, etc. These operators could incorporate computational mechanisms to improve several important algorithm characteristics including population diversity and searching capacities.

Although PSO and ABC are the most popular swarm algorithms for solving complex optimization problems, they present serious flaws such as premature convergence and difficulty to overcome local minima [10,11]. The cause for such problems is associated to the operators that modify individual positions. In such algorithms, during their evolution, the position of each agent for the next iteration is updated yielding an attraction towards the position of the best particle seen so-far (in case of PSO) or towards other randomly chosen individuals (in case of ABC). As the algorithm evolves, those behaviors cause that the entire population concentrates around the best particle or diverges without control. It does favors the premature convergence or damage the exploration-exploitation balance [12,13].

The interesting and exotic collective behavior of social insects have fascinated and attracted researchers for many years. The collaborative swarming behavior observed in these groups provides survival advantages, where insect aggregations of relatively simple and "unintelligent" individuals can accomplish very complex tasks using only limited local information and simple rules of behavior [14]. Social-spiders are a representative example of social insects [15]. A social-spider is a spider species whose members maintain a set of complex cooperative behaviors [16]. Whereas most spiders are solitary and even aggressive toward other members of their own species, social-spiders show a tendency to live in groups, forming long-lasting aggregations often referred to as colonies [17]. In a social-spider colony, each member, depending on its gender, executes a variety of tasks such as predation, mating, web design, and social interaction [17,18]. The web it is an important part of the colony because it is not only used as a common environment for all members, but also as a communication channel among them [19] Therefore, important information (such as trapped prays or mating possibilities) is transmitted by small vibrations through the web. Such information, considered as a local knowledge, is employed by each member to conduct its own cooperative behavior, influencing simultaneously the social regulation of the colony [20].

In this paper, a novel swarm algorithm, called the Social Spider Optimization (SSO) is proposed for solving optimization tasks. The SSO algorithm is based on the simulation of the cooperative behavior of social-spiders. In the proposed algorithm, individuals emulate a group of spiders which interact to each other based on the biological laws of the cooperative colony. The algorithm considers two different search agents (spiders): males and females. Depending on gender, each individual is conducted by a set of different evolutionary operators which mimic different cooperative behaviors that are typical in a colony. Different to most of existent swarm algorithms, in the proposed approach, each individual is modeled considering two genders. Such fact allows not only to emulate in a better realistic way the cooperative behavior of the colony, but also to incorporate computational mechanisms to avoid critical flaws commonly present in the popular PSO and ABC algorithms, such as the premature convergence and the incorrect exploration-exploitation balance. In order to illustrate the proficiency and robustness of the proposed approach, it is compared to other well-known evolutionary methods. The comparison examines several standard benchmark functions which are commonly considered in the literature. The results show a high performance of the proposed method for searching a global optimum in several benchmark functions.

This paper is organized as follows. In Section 2, we introduce basic biological aspects of the algorithm. In Section 3, the novel SSO algorithm and its characteristics are both described. Section 4 presents the experimental results and the comparative study. Finally, in Section 5, conclusions are drawn.

## 2. Biological fundamentals

Social insect societies are complex cooperative systems that self-organize within a set of constraints. Cooperative groups are better at manipulating and exploiting their environment, defending resources and brood, and allowing task specialization among group members [21,22]. A social insect colony functions as an integrated unit that not only possesses the ability to operate at a distributed manner, but also to undertake enormous construction of global projects [23]. It is important to acknowledge that global order in social insects can arise as a result of internal interactions among members.

A few species of spiders have been documented exhibiting a degree of social behavior [15]. The behavior of spiders can be generalized into two basic forms: solitary spiders and social spiders [17]. This classification is made based on the level of cooperative behavior that they exhibit [18]. In one side, solitary spiders create and maintain their own web while live in scarce contact to other individuals of the same species. In contrast, social spiders form colonies that remain together over a communal web with close spatial relationship to other group members [19].

A social spider colony is composed of two fundamental components: its members and the communal web. Members are divided into two different categories: males and females. An interesting characteristic of social-spiders is the highly female-biased population. Some studies suggest that the number of male spiders barely reaches the 30% of the total colony members [17,24]. In the colony, each member, depending on its gender, cooperate in different activities such as building and maintaining the communal web, prey capturing, mating and social contact [20]. Interactions among members are either direct or indirect [25]. Direct interactions imply body contact or the exchange of fluids such as mating. For indirect interactions, the communal web is used as a "medium of communication" which conveys important information that is available to each colony member [19]. This information encoded as small vibrations is a critical aspect for the collective coordination among members [20]. Vibrations are employed by the colony members to decode several messages such as the size of the trapped preys, characteristics of the neighboring members, etc. The intensity of such vibrations depend on the weight and distance of the spiders that have produced them.

In spite of the complexity, all the cooperative global patterns in the colony level are generated as a result of internal interactions among colony members [26]. Such internal interactions involve a set of simple behavioral rules followed by each spider in the colony. Behavioral rules are divided into two different classes: social interaction (cooperative behavior) and mating [27].

As a social insect, spiders perform cooperative interaction with other colony members. The way in which this behavior takes place depends on the spider gender. Female spiders which show a major tendency to socialize present an attraction or dislike over others, irrespectively of gender [17]. For a particular female spider, such attraction or dislike is commonly developed over other spiders according to their vibrations which are emitted over the communal web and represent strong colony members [20]. Since the vibrations depend on the weight and distance of the members which provoke them, stronger vibrations are produced either by big spiders or neighboring members [19]. The bigger a spider is, the better it is considered as a colony member. The final decision of attraction or dislike over a determined member is taken according to an internal state which is influenced by several factors such as reproduction cycle, curiosity and other random phenomena [20].

Different to female spiders, the behavior of male members is reproductive-oriented [28]. Male spiders recognize themselves as a subgroup of alpha males which dominate the colony resources. Therefore, the male population is divided into two classes: dominant and non-dominant male spiders [28]. Dominant male spiders have better fitness characteristics (normally size) in comparison to non-dominant. In a typical behavior, dominant males are attracted to the closest female spider in the communal web. In contrast, non-dominant male spiders tend to concentrate upon the center of the male population as a strategy to take advantage of the resources wasted by dominant males [29].

Mating is an important operation that no only assures the colony survival, but also allows the information exchange among members. Mating in a social-spider colony is performed by dominant males and female members [30]. Under such circumstances, when a dominant male spider locates one or more female members within a specific range, it mates with all the females in order to produce offspring [31].

## 3. The Social Spider Optimization (SSO) algorithm

In this paper, the operational principles from the social-spider colony have been used as guidelines for developing a new swarm optimization algorithm. The SSO assumes that entire search space is a communal web, where all the social-spiders interact to each other. In the proposed approach, each solution within the search space represents a spider position in the communal web. Every spider receives a weight according to the fitness value of the solution that is symbolized by the social-spider. The algorithm models two different search agents (spiders): males and females. Depending on gender, each individual is conducted by a set of different evolutionary operators which mimic different cooperative behaviors that are commonly assumed within the colony.

An interesting characteristic of social-spiders is the highly female-biased populations. In order to emulate this fact, the algorithm starts by defining the number of female and male spiders that will be characterized as individuals in the search space. The number of females $N_f$ is randomly selected within the range of 65% – 90% of the entire population $N$. Therefore, $N_f$ is calculated by the following equation:

$$N_f = \text{floor}\left[(0.9 - \text{rand} \cdot 0.25) \cdot N\right] \tag{1}$$

where rand is a random number between [0,1] whereas floor(·) maps a real number to an integer number. The number of male spiders $N_m$ is computed as the complement between $N$ and $N_f$. It is calculated as follows:

$$N_m = N - N_f \tag{2}$$

Therefore, the complete population **S**, composed by $N$ elements, is divided in two sub-groups **F** and **M**. The Group **F** assembles the set of female individuals ($\mathbf{F} = \{\mathbf{f}_1, \mathbf{f}_2, \ldots, \mathbf{f}_{N_f}\}$) whereas **M** groups the male members ($\mathbf{M} = \{\mathbf{m}_1, \mathbf{m}_2, \ldots, \mathbf{m}_{N_m}\}$), where $\mathbf{S} = \mathbf{F} \cup \mathbf{M}$ ($\mathbf{S} = \{\mathbf{s}_1, \mathbf{s}_2, \ldots, \mathbf{s}_N\}$), such that $\mathbf{S} = \{\mathbf{s}_1 = \mathbf{f}_1, \mathbf{s}_2 = \mathbf{f}_2, \ldots, \mathbf{s}_{N_f} = \mathbf{f}_{N_f}, \mathbf{s}_{N_f+1} = \mathbf{m}_1, \mathbf{s}_{N_f+2} = \mathbf{m}_2, \ldots, \mathbf{s}_N = \mathbf{m}_{N_m}\}$.

*3.1.1 Fitness assignment*

In the biological metaphor, the spider size is the characteristic that evaluates the individual capacity to perform better over its assigned tasks. In the proposed approach, every individual (spider) receives a weight $w_i$ which represents the solution quality that corresponds to the spider $i$ (irrespective of gender) of the population **S**. In order to calculate the weight of every spider the next equation is used:

$$w_i = \frac{J(\mathbf{s}_i) - worst_\mathbf{S}}{best_\mathbf{S} - worst_\mathbf{S}} \tag{3}$$

where $J(\mathbf{s}_i)$ is the fitness value obtained by the evaluation of the spider position $\mathbf{s}_i$ with regard to the objective function $J(\cdot)$. The values $worst_\mathbf{S}$ and $best_\mathbf{S}$ are defined as follows (considering a maximization problem):

$$best_\mathbf{S} = \max_{k \in \{1,2,\ldots,N\}}(J(\mathbf{s}_k)) \text{ and } worst_\mathbf{S} = \min_{k \in \{1,2,\ldots,N\}}(J(\mathbf{s}_k)) \tag{4}$$

*3.1.2 Modeling of the vibrations through the communal web*

The communal web is used as a mechanism to transmit information among the colony members. This information is encoded as small vibrations that are critical for the collective coordination of all individuals in the population. The vibrations depend on the weight and distance of the spider which has generated them. Since the distance is relative to the individual that provokes the vibrations and the member who detects them, members located near to the individual that provokes the vibrations, perceive stronger vibrations in comparison with members located in distant positions. In order to reproduce this process, the vibrations perceived by the individual $i$ as a result of the information transmitted by the member $j$ are modeled according to the following equation:

$$Vib_{i,j} = w_j \cdot e^{-d_{i,j}^2} \tag{5}$$

where the $d_{i,j}$ is the Euclidian distance between the spiders $i$ and $j$, such that $d_{i,j} = \|\mathbf{s}_i - \mathbf{s}_j\|$.

Although it is virtually possible to compute perceived-vibrations by considering any pair of individuals, three special relationships are considered within the SSO approach:

1. Vibrations $Vibc_i$ are perceived by the individual $i$ ($\mathbf{s}_i$) as a result of the information transmitted by the member $c$ ($\mathbf{s}_c$) who is an individual that has two important characteristics: it is the nearest member to $i$ and possesses a higher weight in comparison to $i$ ($w_c > w_i$).

$$Vibc_i = w_c \cdot e^{-d_{i,c}^2} \tag{6}$$

2. The vibrations $Vibb_i$ perceived by the individual $i$ as a result of the information transmitted by the member $b$ ($\mathbf{s}_b$), with $b$ being the individual holding the best weight (best fitness value) of the entire population $\mathbf{S}$, such that $w_b = \max_{k \in \{1,2,\ldots,N\}}(w_k)$.

$$Vibb_i = w_b \cdot e^{-d_{i,b}^2} \tag{7}$$

3. The vibrations $Vibf_i$ perceived by the individual $i$ ($\mathbf{s}_i$) as a result of the information transmitted by the member $f$ ($\mathbf{s}_f$), with $f$ being the nearest female individual to $i$.

$$Vibf_i = w_f \cdot e^{-d_{i,f}^2} \tag{7}$$

Fig. 1 shows the configuration of each special relationship: a) $Vibc_i$, b) $Vibb_i$ and c) $Vibf_i$.

*3.1.3 Initializing the population*

Like other evolutionary algorithms, the SSO is an iterative process whose first step is to randomly initialize the entire population (female and male). The algorithm begins by initializing the set $\mathbf{S}$ of $N$ spider positions. Each spider position, $\mathbf{f}_i$ or $\mathbf{m}_i$, is a $n$-dimensional vector containing the parameter values to be optimized. Such values are randomly and uniformly distributed between the pre-specified lower initial parameter bound $p_j^{low}$ and the upper initial parameter bound $p_j^{high}$, just as it described by the following expressions:

$$f_{i,j}^0 = p_j^{low} + \text{rand}(0,1) \cdot (p_j^{high} - p_j^{low}) \qquad m_{k,j}^0 = p_j^{low} + \text{rand}(0,1) \cdot (p_j^{high} - p_j^{low}) \tag{8}$$
$$i = 1, 2, \ldots, N_f; j = 1, 2, \ldots, n \qquad k = 1, 2, \ldots, N_m; j = 1, 2, \ldots, n$$

where $j$, $i$ and $k$ are the parameter and individual indexes respectively whereas zero signals the initial population. The function rand(0,1) generates a random number between 0 and 1. Hence, $f_{i,j}$ is the $j$-th parameter of the $i$-th female spider position.

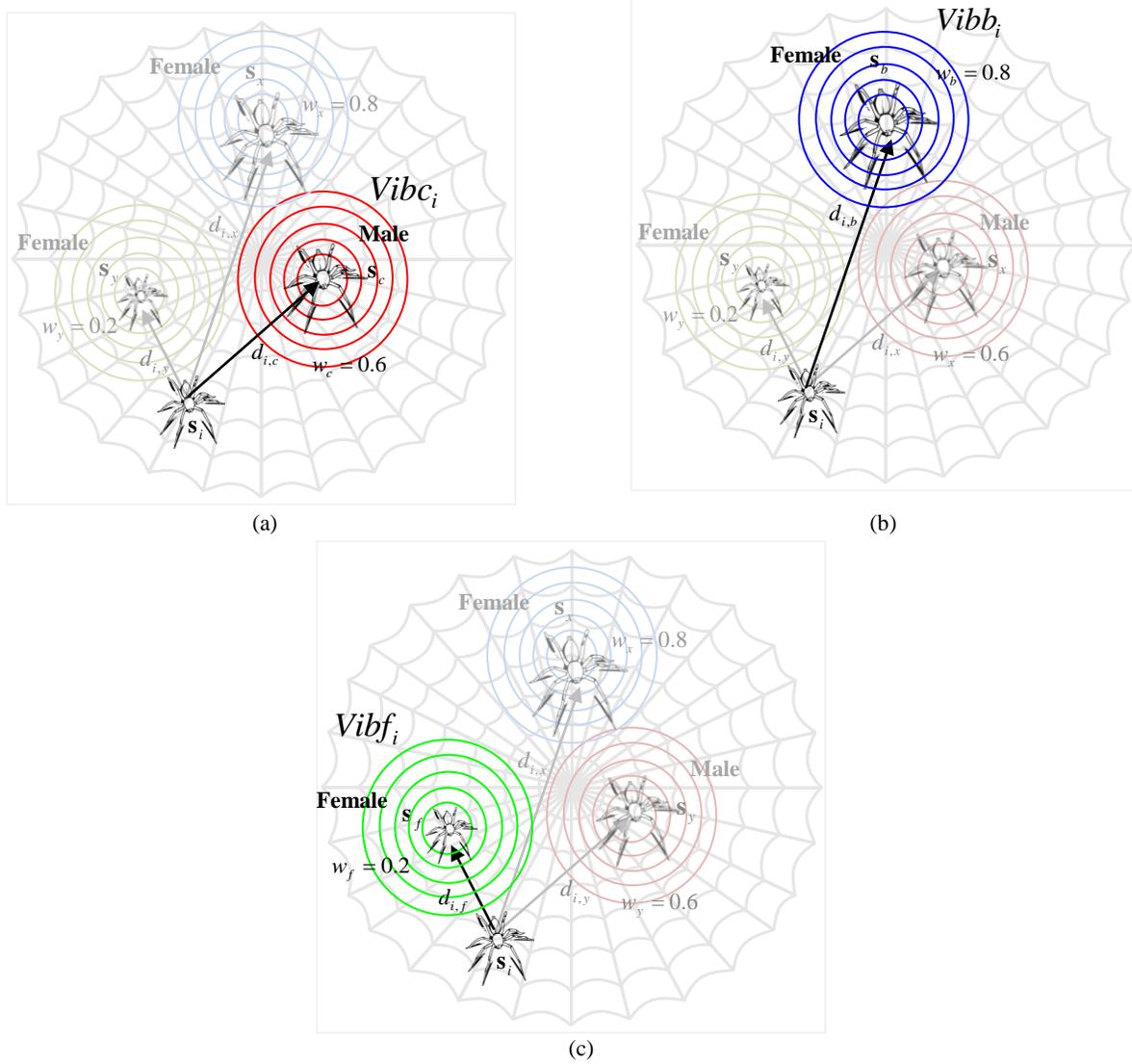

**Fig. 1.** Configuration of each special relation: a) $Vibc_i$, b) $Vibb_i$ and c) $Vibf_i$.

*3.1.4 Cooperative operators*

Female cooperative operator

Social-spiders perform cooperative interaction over other colony members. The way in which this behavior takes place depends on the spider gender. Female spiders present an attraction or dislike over others irrespective of gender. For a particular female spider, such attraction or dislike is commonly developed over other spiders according to their vibrations which are emitted over the communal web. Since vibrations depend on the weight and distance of the members which have originated them, strong vibrations are produced either by big spiders or other neighboring members lying nearby the individual which is perceiving them. The final decision of attraction or dislike over a determined member is taken considering an internal state which is influenced by several factors such as reproduction cycle, curiosity and other random phenomena.

In order to emulate the cooperative behavior of the female spider, a new operator is defined. The operator considers the position change of the female spider *i* at each iteration. Such position change, which can be of attraction or repulsion, is computed as a combination of three different elements. The first one involves the change in regard to the nearest member to *i* that holds a higher weight and produces the vibration $Vibc_i$. The second one considers the change regarding the best individual of the entire population **S** who produces the vibration $Vibb_i$. Finally, the third one incorporates a random movement.

Since the final movement of attraction or repulsion depends on several random phenomena, the selection is modeled as a stochastic decision. For this operation, a uniform random number $r_m$ is generated within the range [0,1]. If $r_m$ is smaller than a threshold *PF*, an attraction movement is generated; otherwise, a repulsion movement is produced. Therefore, such operator can be modeled as follows:

$$\mathbf{f}_i^{k+1} = \begin{cases} \mathbf{f}_i^k + \alpha \cdot Vibc_i \cdot (\mathbf{s}_c - \mathbf{f}_i^k) + \beta \cdot Vibb_i \cdot (\mathbf{s}_b - \mathbf{f}_i^k) + \delta \cdot (\text{rand} - \frac{1}{2}) & \text{with probability } PF \\ \mathbf{f}_i^k - \alpha \cdot Vibc_i \cdot (\mathbf{s}_c - \mathbf{f}_i^k) - \beta \cdot Vibb_i \cdot (\mathbf{s}_b - \mathbf{f}_i^k) + \delta \cdot (\text{rand} - \frac{1}{2}) & \text{with probability } 1\text{-}PF \end{cases} \quad (9)$$

where $\alpha$, $\beta$, $\delta$ and rand are random numbers between [0,1] whereas *k* represents the iteration number. The individual $\mathbf{s}_c$ and $\mathbf{s}_b$ represent the nearest member to *i* that holds a higher weight and the best individual of the entire population **S**, respectively.

Under this operation, each particle presents a movement which combines the past position that holds the attraction or repulsion vector over the local best element $\mathbf{s}_c$ and the global best individual $\mathbf{s}_b$ seen so-far. This particular type of interaction avoids the quick concentration of particles at only one point and encourages each particle to search around the local candidate region within its neighborhood ($\mathbf{s}_c$), rather than interacting to a particle ($\mathbf{s}_b$) in a distant region of the domain. The use of this scheme has two advantages. First, it prevents the particles from moving towards the global best position, making the algorithm less susceptible to premature convergence. Second, it encourages particles to explore their own neighborhood thoroughly before converging towards the global best position. Therefore, it provides the algorithm with global search ability and enhances the exploitative behavior of the proposed approach.

Male cooperative operator

According to the biological behavior of the social-spider, male population is divided into two classes: dominant and non-dominant male spiders. Dominant male spiders have better fitness characteristics (usually regarding the size) in comparison to non-dominant. Dominant males are attracted to the closest female spider in the communal web. In contrast, non-dominant male spiders tend to concentrate in the center of the male population as a strategy to take advantage of resources that are wasted by dominant males.

For emulating such cooperative behavior, the male members are divided into two different groups (dominant members **D** and non-dominant members **ND**) according to their position with regard to the median member. Male members, with a weight value above the median value within the male population, are considered the dominant individuals **D**. On the other hand, those under the median value are labeled as non-dominant **ND** males. In order to implement such computation, the male population **M** ( $\mathbf{M} = \{\mathbf{m}_1, \mathbf{m}_2, \ldots, \mathbf{m}_{N_m}\}$ ) is arranged according to their weight value in decreasing order. Thus, the individual whose weight $w_{N_f+m}$ is located in the middle is considered the median male member. Since indexes of the male population **M** in regard to the entire population **S** are increased by the number of female members $N_f$, the median weight is indexed by $N_f+m$. According to this, change of positions for the male spider can be modeled as follows:

$$\mathbf{m}_i^{k+1} = \begin{cases} \mathbf{m}_i^k + \alpha \cdot Vibf_i \cdot (\mathbf{s}_f - \mathbf{m}_i^k) + \delta \cdot (\text{rand} - \frac{1}{2}) & \text{if } w_{N_f+i} > w_{N_f+m} \\ \mathbf{m}_i^k + \alpha \cdot \left( \frac{\sum_{h=1}^{N_m} \mathbf{m}_h^k \cdot w_{N_f+h}}{\sum_{h=1}^{N_m} w_{N_f+h}} - \mathbf{m}_i^k \right) & \text{if } w_{N_f+i} \leq w_{N_f+m} \end{cases}, \qquad (10)$$

where the individual $\mathbf{s}_f$ represents the nearest female individual to the male member $i$ whereas $\left( \sum_{h=1}^{N_m} \mathbf{m}_h^k \cdot w_{N_f+h} / \sum_{h=1}^{N_m} w_{N_f+h} \right)$ correspond to the weighted mean of the male population **M**.

By using this operator, two different behaviors are produced. First, the set **D** of particles is attracted to others in order to provoke mating. Such behavior allows incorporating diversity into the population. Second, the set **ND** of particles is attracted to the weighted mean of the male population **M.** This fact is used to partially control the search process according to the average performance of a sub-group of the population. Such mechanism acts as a filter which avoids that very good individuals or extremely bad individuals influence the search process.

*3.1.5 Mating operator*

Mating in a social-spider colony is performed by dominant males and the female members. Under such circumstances, when a dominant male $\mathbf{m}_g$ spider ($g \in \mathbf{D}$) locates a set $\mathbf{E}^g$ of female members within a specific range *r* (range of mating), it mates, forming a new brood $\mathbf{s}_{new}$ which is generated considering all the elements of the set $\mathbf{T}^g$ that, in turn, has been generated by the union $\mathbf{E}^g \cup \mathbf{m}_g$. It is important to emphasize that if the set $\mathbf{E}^g$ is empty, the mating operation is canceled. The range *r* is defined as a radius which depends on the size of the search space. Such radius *r* is computed according to the following model:

$$r = \frac{\sum_{j=1}^{n}(p_j^{high} - p_j^{low})}{2 \cdot n} \qquad (10)$$

In the mating process, the weight of each involved spider (elements of $\mathbf{T}^g$) defines the probability of influence for each individual into the new brood. The spiders holding a heavier weight are more likely to influence the new product, while elements with lighter weight have a lower probability. The influence probability $Ps_i$ of each member is assigned by the roulette method, which is defined as follows:

$$Ps_i = \frac{w_i}{\sum_{j \in \mathbf{T}^k} w_j}, \qquad (11)$$

where $i \in \mathbf{T}^g$.

Once the new spider is formed, it is compared to the new spider candidate $\mathbf{s}_{new}$ holding the worst spider $\mathbf{s}_{wo}$ of the colony, according to their weight values (where $w_{wo} = \min_{l \in \{1,2,\ldots,N\}}(w_l)$). If the new spider is better than the worst spider, the worst spider is replaced by the new one. Otherwise, the new spider is discarded and the population does not suffer changes. In case of replacement, the new spider assumes the gender and index from the replaced spider. Such fact assures that the entire population $\mathbf{S}$ maintains the original rate between female and male members.

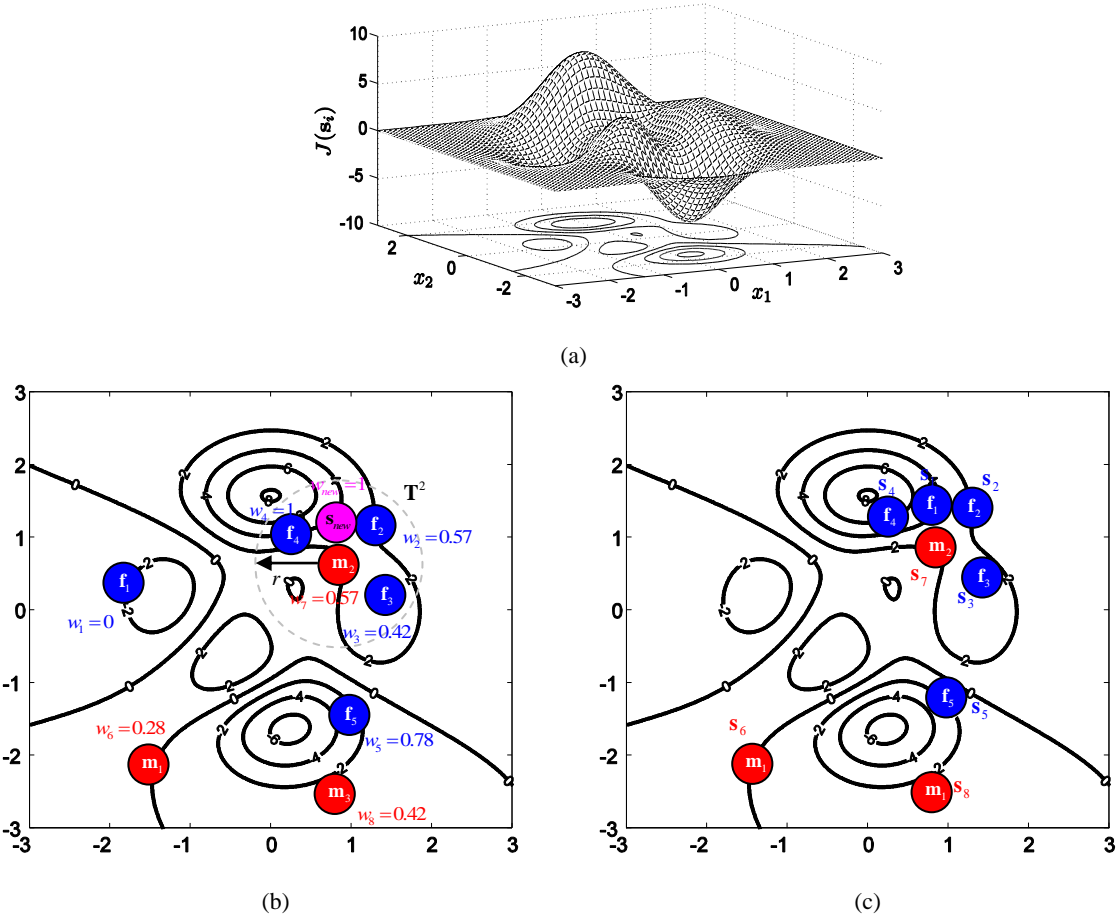

**Fig. 2.** Example of the mating operation: (a) optimization problem, (b) initial configuration before mating and (c) configuration after the mating operation.

In order to demonstrate the mating operation, Fig. 2a illustrates a simple optimization problem. As an example, it is assumed a population $\mathbf{S}$ of eight different 2-dimensional members ($N$=8), five females ($N_f = 5$) and three males ($N_m = 3$). Fig. 2b shows the initial configuration of the proposed example with three different female members $\mathbf{f}_2(\mathbf{s}_2), \mathbf{f}_3(\mathbf{s}_3)$ and $\mathbf{f}_4(\mathbf{s}_4)$ constituting the set $\mathbf{E}^2$ which is located inside of the influence range $r$ of a dominant male $\mathbf{m}_2(\mathbf{s}_7)$. Then, the new candidate spider $\mathbf{s}_{new}$ is generated from the elements $\mathbf{f}_2, \mathbf{f}_3, \mathbf{f}_4$ and $\mathbf{m}_2$ which constitute the set $\mathbf{T}^2$. Therefore, the value of the first decision variable $s_{new,1}$ for the new spider is chosen by means of the roulette mechanism considering the values already existing from the set $\{f_{2,1}, f_{3,1}, f_{4,1}, m_{2,1}\}$. The value of the second decision variable $s_{new,2}$ is also chosen in the same manner. Table 1 shows the data for constructing the new spider through the roulette method. Once the new spider $s_{new}$ is formed, its weight $w_{new}$ is calculated. As $s_{new}$ is better than the worst member $f_1$ that is present in the population **S**, $f_1$ is replaced by $s_{new}$. Therefore, $s_{new}$ assumes the same gender and index from $f_1$. Fig. 2c shows the configuration of **S** after the mating process.

Under this operation, new generated particles locally exploit the search space inside the mating range in order to find better individuals.

| Spider | | Position | $w_i$ | $Ps_i$ | Roulette |
|---|---|---|---|---|---|
| $s_1$ | $f_1$ | (-1.9,0.3) | 0.00 | - | |
| $s_2$ | $f_2$ | (1.4,1.1) | 0.57 | 0.22 | |
| $s_3$ | $f_3$ | (1.5,0.2) | 0.42 | 0.16 | |
| $s_4$ | $f_4$ | (0.4,1.0) | 1.00 | 0.39 | |
| $s_5$ | $f_5$ | (1.0,-1.5) | 0.78 | - | |
| $s_6$ | $m_1$ | (-1.3,-1.9) | 0.28 | - | |
| $s_7$ | $m_2$ | (0.9,0.7) | 0.57 | 0.22 | |
| $s_8$ | $m_3$ | (0.8,-2.6) | 0.42 | - | |
| $s_{new}$ | | (0.9,1.1) | 1.00 | - | |

Roulette: $f_2$ 25%, $f_4$ 41%, $f_3$ 17%, $m_3$ 17%

**Table 1.** Data for constructing the new spider $s_{new}$ through the roulette method.

### 3.1.6. Computational procedure

The computational procedure for the proposed algorithm can be summarized as follows:

| Step 1: | Considering $N$ as the total number of $n$-dimensional colony members, define the number of male $N_m$ and females $N_f$ spiders in the entire population **S**. |
|---|---|

$$N_f = \text{floor}\left[(0.9 - \text{rand} \cdot 0.25) \cdot N\right] \text{ and } N_m = N - N_f,$$

where rand is a random number between [0,1] whereas floor(·) maps a real number to an integer number.

| Step 2: | Initialize randomly the female ($\mathbf{F} = \{f_1, f_2, \ldots, f_{N_f}\}$) and male ($\mathbf{M} = \{m_1, m_2, \ldots, m_{N_m}\}$) members (where $\mathbf{S} = \{s_1 = f_1, s_2 = f_2, \ldots, s_{N_f} = f_{N_f}, s_{N_f+1} = m_1, s_{N_f+2} = m_2, \ldots, s_N = m_{N_m}\}$ and calculate the radius of mating. |
|---|---|

$$r = \frac{\sum_{j=1}^{n}(p_j^{high} - p_j^{low})}{2 \cdot n}$$

```
for (i=1; i< N_f +1; i++)
  for(j=1; j<n+1; j++)
    f_{i,j}^0 = p_j^{low} + rand(0,1) \cdot (p_j^{high} - p_j^{low})
  end for
end for
for (k=1; k< N_m +1; k++)
  for(j=1; j<n+1; j++)
    m_{k,j}^0 = p_j^{low} + rand \cdot (p_j^{high} - p_j^{low})
  end for
end for
```

**Step 3:** Calculate the weight of every spider of **S** (section 3.1.1).
for ($i=1, i<N+1; i++$)

$$w_i = \frac{J(\mathbf{s}_i) - worst_{\mathbf{S}}}{best_{\mathbf{S}} - worst_{\mathbf{S}}}$$

where $best_{\mathbf{S}} = \max_{k \in \{1,2,\ldots,N\}}(J(\mathbf{s}_k))$ and $worst_{\mathbf{S}} = \min_{k \in \{1,2,\ldots,N\}}(J(\mathbf{s}_k))$

end for

**Step 4:** Move female spiders according to the female cooperative operator (section 3.1.4).
for ($i=1; i< N_f +1; i++$)

Calculate $Vibc_i$ and $Vibb_i$ (Section 3.1.2)

If ($r_m < PF$); where $r_m \in \text{rand}(0,1)$

$$\mathbf{f}_i^{k+1} = \mathbf{f}_i^k + \alpha \cdot Vibc_i \cdot (\mathbf{s}_c - \mathbf{f}_i^k) + \beta \cdot Vibb_i \cdot (\mathbf{s}_b - \mathbf{f}_i^k) + \delta \cdot (\text{rand} - \frac{1}{2})$$

else if

$$\mathbf{f}_i^{k+1} = \mathbf{f}_i^k - \alpha \cdot Vibc_i \cdot (\mathbf{s}_c - \mathbf{f}_i^k) - \beta \cdot Vibb_i \cdot (\mathbf{s}_b - \mathbf{f}_i^k) + \delta \cdot (\text{rand} - \frac{1}{2})$$

end if
end for

**Step 5:** Move the male spiders according to the male cooperative operator (section 3.1.4).
Find the median male individual ($w_{N_f+m}$) from **M**.

for ($i=1; i< N_m +1; i++$)

Calculate $Vibf_i$ (section 3.1.2)

If ($w_{N_f+i} > w_{N_f+m}$)

$$\mathbf{m}_i^{k+1} = \mathbf{m}_i^k + \alpha \cdot Vibf_i \cdot (\mathbf{s}_f - \mathbf{m}_i^k) + \delta \cdot (\text{rand} - \frac{1}{2})$$

Else if

$$\mathbf{m}_i^{k+1} = \mathbf{m}_i^k + \alpha \cdot \left( \frac{\sum_{h=1}^{N_m} \mathbf{m}_h^k \cdot w_{N_f+h}}{\sum_{h=1}^{N_m} w_{N_f+h}} - \mathbf{m}_i^k \right)$$

end if
end for

**Step 6:** Perform the mating operation (Section 3.1.5).
for ($i=1; i< N_m +1; i++$)

If ($\mathbf{m}_i \in \mathbf{D}$)

Find $\mathbf{E}^i$

If ($\mathbf{E}^i$ is not empty)

Form $\mathbf{s}_{new}$ using the roulette method

If ($w_{new} > w_{wo}$)

$\mathbf{s}_{wo} = \mathbf{s}_{new}$

end if
end if
end if
end for

**Step 7:** If the stop criteria is met, the process is finished; otherwise, go back to Step 3

*3.1.7. Discussion about the SSO algorithm*

Evolutionary algorithms (EA) have been widely employed for solving complex optimization problems. These methods are found to be more powerful than conventional methods based on formal logics or mathematical programming [32]. In an EA algorithm, search agents have to decide whether to explore unknown search positions or to exploit already tested positions in order to improve their solution quality. Pure exploration degrades the precision of the evolutionary process but increases its capacity to find new potential solutions. On the other hand, pure exploitation allows refining existent solutions but adversely drives the process to local optimal solutions. Therefore, the ability of an EA to find a global optimal solutions depends on its capacity to find a good balance between the exploitation of found-so-far elements and the exploration of the search space [33]. So far, the exploration–exploitation dilemma has been an unsolved issue within the framework of evolutionary algorithms.

EA defines individuals with the same property, performing virtually the same behavior. Under these circumstances, algorithms waste the possibility to add new and selective operators as a result of considering individuals with different characteristics. These operators could incorporate computational mechanisms to improve several important algorithm characteristics such as population diversity or searching capacities.

On the other hand, PSO and ABC are the most popular swarm algorithms for solving complex optimization problems. However, they present serious flaws such as premature convergence and difficulty to overcome local minima [10,11]. Such problems arise from operators that modify individual positions. In such algorithms, the position of each agent in the next iteration is updated yielding an attraction towards the position of the best particle seen so-far (in case of PSO) or any other randomly chosen individual (in case of ABC). Such behaviors produce that the entire population concentrates around the best particle or diverges without control as the algorithm evolves, either favoring the premature convergence or damaging the exploration-exploitation balance [12,13].

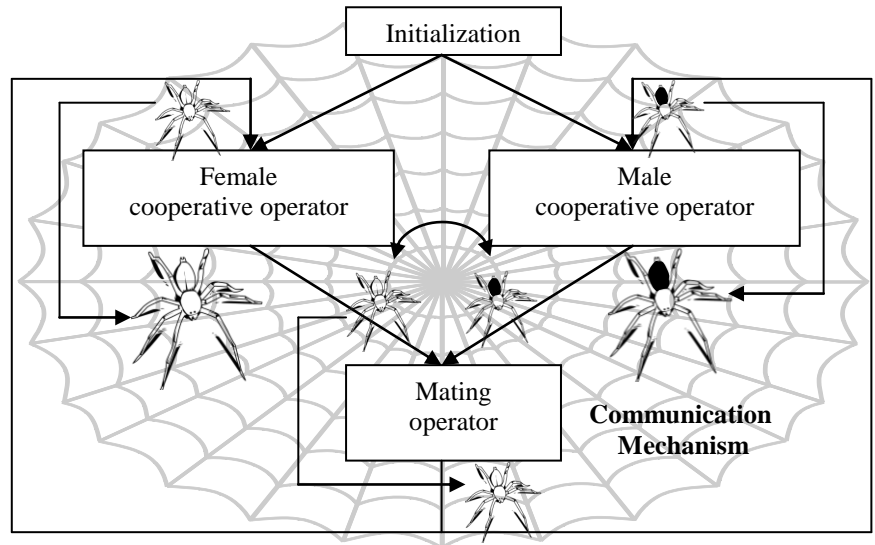

**Fig. 3.** Schematic representation of the SSO algorithm-data-flow

Different to other EA, at SSO each individual is modeled considering the gender. Such fact allows incorporating computational mechanisms to avoid critical flaws such as premature convergence and incorrect exploration-exploitation balance commonly present in both, the PSO and the ABC algorithm. From an optimization point of view, the use of the social-spider behavior as a metaphor introduces interesting concepts in EA: the fact of dividing the entire population into different search-agent categories and the employment of specialized operators that are applied selectively to each of them. By using this framework, it is possible to improve the balance between exploitation and exploration, yet preserving the same population, i.e. individuals who have achieved efficient exploration (female spiders) and individuals that verify extensive exploitation (male spiders). Furthermore, the social-spider behavior mechanism introduces an interesting computational scheme with three important particularities: first, individuals are separately processed according to their characteristics. Second, operators share the same communication mechanism allowing the employment of important information of the evolutionary process to modify the influence of each operator. Third, although operators modify the position of only an individual type, they use global information (positions of all individual types) in order to perform such modification. Fig. 3 presents a schematic representation of the algorithm-data-flow. According to Fig. 3, the female cooperative and male cooperative operators process only female or male individuals, respectively. However, the mating operator modifies both individual types.

## 4. Experimental results

A comprehensive set of 19 functions, which have been collected from Refs. [34-40], has been used to test the performance of the proposed approach. Table A1 in the Appendix A presents the benchmark functions used in our experimental study. In the table, *n* indicates the function dimension, $f(\mathbf{x}^*)$ the optimum value of the function, $\mathbf{x}^*$ the optimum position and $S$ the search space (subset of $R^n$). A detailed description of each function is given in the Appendix A.

*4.1 Performance comparison to other swarm algorithms*

We have applied the SSO algorithm to 19 functions whose results have been compared to those produced by the Particle Swarm Optimization (PSO) method [3] and the Artificial Bee Colony (ABC) algorithm [4]. These are considered as the most popular swarm algorithms for many optimization applications. In all comparisons, the population has been set to 50 individuals. The maximum iteration number for all functions has been set to 1000. Such stop criterion has been selected to maintain compatibility to similar works reported in the literature [41,42].

The parameter setting for each algorithm in the comparison is described as follows:

1. PSO: The parameters are set to $c_1 = 2$ and $c_2 = 2$; besides, the weight factor decreases linearly from 0.9 to 0.2 [3].
2. ABC: The algorithm has been implemented using the guidelines provided by its own reference [4], using the parameter *limit*=100.
3. SSO: Once it has been determined experimentally, the parameter *PF* has been set to 0.7. It is kept for all experiments in this section.

The experiment compares the SSO to other algorithms such as PSO and ABC. The results for 30 runs are reported in Table 2 considering the following performance indexes: the Average Best-so-far (AB) solution, the Median Best-so-far (MB) and the Standard Deviation (SD) of best-so-far solution. The best outcome for each function is boldfaced. According to this table, SSO delivers better results than PSO and ABC for all functions. In particular, the test remarks the largest difference in performance which is directly related to a better trade-off between exploration and exploitation.

Fig. 4 presents the evolution curves for PSO, ABC and the proposed algorithm considering as examples the functions $f_1$, $f_3$, $f_5$, $f_{10}$, $f_{15}$ and $f_{19}$ from the experimental set. Among them, the rate of convergence of SSO is the fastest, which finds the best solution in less of 400 iterations on average while the other three algorithms need much more iterations.

A non-parametric statistical significance proof known as the Wilcoxon's rank sum test for independent samples [43,44] has been conducted over the "average best-so-far" (AB) data of Table 2, with an 5% significance level. Table 3 reports the *p*-values produced by Wilcoxon's test for the pair-wise comparison of the "average best so-far" of two groups. Such groups are constituted by SSO vs. PSO and SSO vs. ABC. As a null hypothesis, it is assumed that there is no significant difference between mean values of the two algorithms. The alternative hypothesis considers a significant difference between the "average best-so-far" values of both approaches. All *p*-values reported in Table 3 are less than 0.05 (5% significance level) which is a strong evidence against the null hypothesis. Therefore, such evidence indicates that SSO results are statistically significant and it has not occurred by coincidence (i.e. due to common noise contained in the process).

|  |  | SSO | ABC | PSO |
|---|---|---|---|---|
| $f_1(x)$ | AB | **1.96E-03** | 2.90E-03 | 1.00E+03 |
|  | MB | 2.81E-03 | 1.50E-03 | **2.08E-09** |
|  | SD | 9.96E-04 | 1.44E-03 | 3.05E+03 |
| $f_2(x)$ | AB | **1.37E-02** | 1.35E-01 | 5.17E+01 |
|  | MB | **1.34E-02** | 1.05E-01 | 5.00E+01 |
|  | SD | **3.11E-03** | 8.01E-02 | 2.02E+01 |
| $f_3(x)$ | AB | **4.27E-02** | 1.13E+00 | 8.63E+04 |
|  | MB | **3.49E-02** | 6.11E-01 | 8.00E+04 |
|  | SD | **3.11E-02** | 1.57E+00 | 5.56E+04 |
| $f_4(x)$ | AB | **5.40E-02** | 5.82E+01 | 1.47E+01 |
|  | MB | **5.43E-02** | 5.92E+01 | 1.51E+01 |
|  | SD | **1.01E-02** | 7.02E+00 | 3.13E+00 |
| $f_5(x)$ | AB | **1.14E+02** | 1.38E+02 | 3.34E+04 |
|  | MB | **5.86E+01** | 1.32E+02 | 4.03E+02 |
|  | SD | **3.90E+01** | 1.55E+02 | 4.38E+04 |
| $f_6(x)$ | AB | **2.68E-03** | 4.06E-03 | 1.00E+03 |
|  | MB | 2.68E-03 | 3.74E-03 | **1.66E-09** |
|  | SD | **6.05E-04** | 2.98E-03 | 3.06E+03 |
| $f_7(x)$ | AB | **1.20E+01** | 1.21E+01 | 1.50E+01 |
|  | MB | **1.20E+01** | 1.23E+01 | 1.37E+01 |
|  | SD | **5.76E-01** | 9.00E-01 | 4.75E+00 |
| $f_8(x)$ | AB | **2.14E+00** | 3.60E+00 | 3.12E+04 |
|  | MB | 3.64E+00 | 8.04E-01 | 2.08E+02 |
|  | SD | **1.26E+00** | 3.54E+00 | 5.74E+04 |
| $f_9(x)$ | AB | **6.92E-05** | 1.44E-04 | 2.47E+00 |
|  | MB | **6.80E-05** | 8.09E-05 | 9.09E-01 |
|  | SD | **4.02E-05** | 1.69E-04 | 3.27E+00 |
| $f_{10}(x)$ | AB | **4.44E-04** | 1.10E-01 | 6.93E+02 |
|  | MB | **4.05E-04** | 4.97E-02 | 5.50E+02 |
|  | SD | **2.90E-04** | 1.98E-01 | 6.48E+02 |
| $f_{11}(x)$ | AB | **6.81E+01** | 3.12E+02 | 4.11E+02 |
|  | MB | **6.12E+01** | 3.13E+02 | 4.31E+02 |
|  | SD | **3.00E+01** | 4.31E+01 | 1.56E+02 |
| $f_{12}(x)$ | AB | **5.39E-05** | 1.18E-04 | 4.27E+07 |
|  | MB | **5.40E-05** | 1.05E-04 | 1.04E-01 |
|  | SD | **1.84E-05** | 8.88E-05 | 9.70E+07 |
| $f_{13}(x)$ | AB | **1.76E-03** | 1.87E-03 | 5.74E-01 |
|  | MB | 1.12E-03 | 1.69E-03 | **1.08E-05** |
|  | SD | **6.75E-04** | 1.47E-03 | 2.36E+00 |
| $f_{14}(x)$ | AB | -9.36E+02 | **-9.69E+02** | -9.63E+02 |
|  | MB | -9.36E+02 | -9.60E+02 | **-9.92E+02** |
|  | SD | **1.61E+01** | 6.55E+01 | 6.66E+01 |
| $f_{15}(x)$ | AB | **8.59E+00** | 2.64E+01 | 1.35E+02 |
|  | MB | **8.78E+00** | 2.24E+01 | 1.36E+02 |
|  | SD | **1.11E+00** | 1.06E+01 | 3.73E+01 |
| $f_{16}(x)$ | AB | **1.36E-02** | 6.53E-01 | 1.14E+01 |
|  | MB | **1.39E-02** | 6.39E-01 | 1.43E+01 |
|  | SD | **2.36E-03** | 3.09E-01 | 8.86E+00 |
| $f_{17}(x)$ | AB | **3.29E-03** | 5.22E-02 | 1.20E+01 |
|  | MB | 3.21E-03 | 4.60E-02 | **1.35E-02** |
|  | SD | **5.49E-04** | 3.42E-02 | 3.12E+01 |
| $f_{18}(x)$ | AB | **1.87E+00** | 2.13E+00 | 1.26E+03 |
|  | MB | **1.61E+00** | 2.14E+00 | 5.67E+02 |
|   | SD | **1.20E+00** | 1.22E+00 | 1.12E+03 |
|   | AB | **2.74E-01** | 4.14E+00 | 1.53E+00 |
| $f_{19}(x)$ | MB | **3.00E-01** | 4.10E+00 | 5.50E-01 |
|   | SD | **5.17E-02** | 4.69E-01 | 2.94E+00 |

**Table 2.** Minimization results of benchmark functions of Table A with $n$=30. Maximum number of iterations=1000.

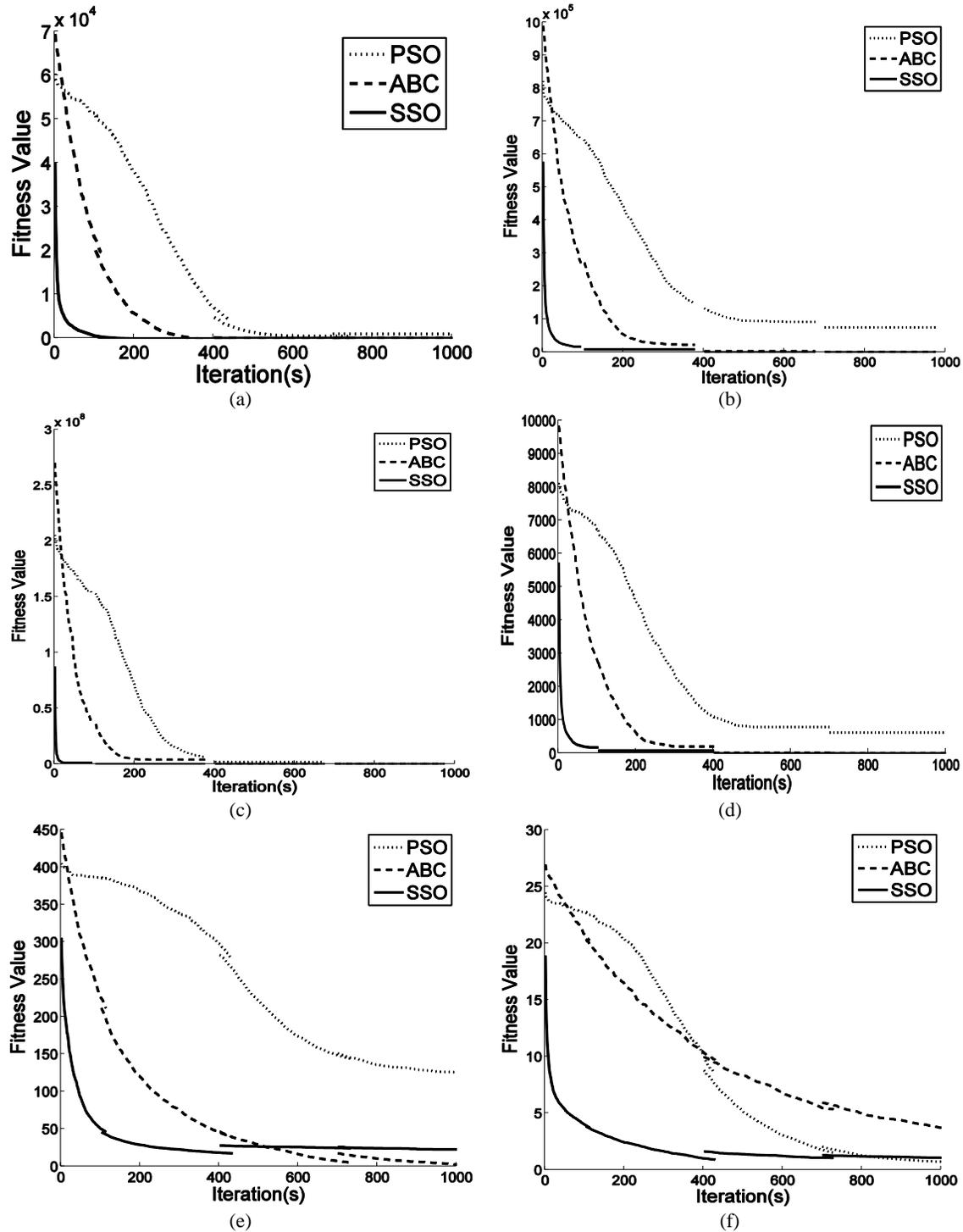

(a)

(b)

(c)

(d)

(e)

(f)

**Fig. 4.** Evolution curves for PSO, ABC and the proposed algorithm considering as examples the functions (a) $f_1$, (b) $f_3$, (c) $f_5$, (d) $f_{10}$, (e) $f_{15}$ and (f) $f_{19}$ from the experimental set.

| Function | SSO vs ABC | SSO vs PSO |
|---|---|---|
| $f_1(x)$ | 0.041 | 1.8E-05 |
| $f_2(x)$ | 0.048 | 0.059 |
| $f_3(x)$ | 5.4E-04 | 6.2E-07 |
| $f_4(x)$ | 1.4E-07 | 4.7E-05 |
| $f_5(x)$ | 0.045 | 7.1E-07 |
| $f_6(x)$ | 2.3E-04 | 5.5E-08 |
| $f_7(x)$ | 0.048 | 0.011 |
| $f_8(x)$ | 0.017 | 0.043 |
| $f_9(x)$ | 8.1E-04 | 2.5E-08 |
| $f_{10}(x)$ | 4.6E-06 | 1.7E-09 |
| $f_{11}(x)$ | 9.2E-05 | 7.8E-06 |
| $f_{12}(x)$ | 0.022 | 1.1E-10 |
| $f_{13}(x)$ | 0.048 | 2.6E-05 |
| $f_{14}(x)$ | 0.044 | 0.049 |
| $f_{15}(x)$ | 4.5E-05 | 7.9E-08 |
| $f_{16}(x)$ | 2.8E-05 | 4.1E-06 |
| $f_{17}(x)$ | 7.1E-04 | 6.2E-10 |
| $f_{18}(x)$ | 0.013 | 8.3E-10 |
| $f_{19}(x)$ | 4.9E-05 | 5.1E-08 |

**Table 3.** *p*-values produced by Wilcoxon's test comparing SSO vs. ABC and SSO vs. PSO, over the "average best-so-far" (AB) values from Table 2.

## 5. Conclusions

In this paper, a novel swarm algorithm called the Social Spider Optimization (SSO) has been proposed for solving optimization tasks. The SSO algorithm is based on the simulation of the cooperative behavior of social-spiders whose individuals emulate a group of spiders which interact to each other based on the biological laws of a cooperative colony. The algorithm considers two different search agents (spiders): male and female. Depending on gender, each individual is conducted by a set of different evolutionary operators which mimic different cooperative behaviors within the colony.

In contrast to most of existent swarm algorithms, the proposed approach models each individual considering two genders. Such fact allows not only to emulate the cooperative behavior of the colony in a realistic way, but also to incorporate computational mechanisms to avoid critical flaws commonly delivered by the popular

PSO and ABC algorithms, such as the premature convergence and the incorrect exploration-exploitation balance.

SSO has been experimentally tested considering a suite of 19 benchmark functions. The performance of SSO has been also compared to the following swarm algorithms: the Particle Swarm Optimization method (PSO) [16], and the Artificial Bee Colony (ABC) algorithm [38]. Results have confirmed a acceptable performance of the proposed method in terms of the solution quality of the solution for all tested benchmark functions.

The SSO's remarkable performance is associated with two different reasons: (i) their operators allow a better particle distribution in the search space, increasing the algorithm's ability to find the global optima; and (ii) the division of the population into different individual types, provides the use of different rates between exploration and exploitation during the evolution process.

**Appendix A. List of benchmark functions**

| Name | Function | $S$ | Dim | Minimum |
|---|---|---|---|---|
| Sphere | $f_1(\mathbf{x}) = \sum_{i=1}^{n} x_i^2$ | $[-100,100]^n$ | $n=30$ | $\mathbf{x}^* = (0,\ldots,0);$ $f(\mathbf{x}^*) = 0$ |
| Schwefel 2.22 | $f_2(\mathbf{x}) = \sum_{i=1}^{n} |x_i| + \prod_{i=1}^{n} |x_i|$ | $[-10,10]^n$ | $n=30$ | $\mathbf{x}^* = (0,\ldots,0);$ $f(\mathbf{x}^*) = 0$ |
| Schwefel 1.2 | $f_3(\mathbf{x}) = \sum_{i=1}^{n} \left( \sum_{j=1}^{i} x_j \right)^2$ | $[-100,100]^n$ | $n=30$ | $\mathbf{x}^* = (0,\ldots,0);$ $f(\mathbf{x}^*) = 0$ |
| F4 | $f_4(\mathbf{x}) = 418.9829n + \sum_{i=1}^{n} \left( -x_i \sin\left(\sqrt{|x_i|}\right) \right)$ | $[-100,100]^n$ | $n=30$ | $\mathbf{x}^* = (0,\ldots,0);$ $f(\mathbf{x}^*) = 0$ |
| Rosenbrock | $f_5(\mathbf{x}) = \sum_{i=1}^{n-1} \left[ 100(x_{i+1} - x_i^2)^2 + (x_i - 1)^2 \right]$ | $[-30,30]^n$ | $n=30$ | $\mathbf{x}^* = (1,\ldots,1);$ $f(\mathbf{x}^*) = 0$ |
| Step | $f_6(\mathbf{x}) = \sum_{i=1}^{n} \left( \lfloor x_i + 0.5 \rfloor \right)^2$ | $[-100,100]^n$ | $n=30$ | $\mathbf{x}^* = (0,\ldots,0);$ $f(\mathbf{x}^*) = 0$ |
| Quartic | $f_7(\mathbf{x}) = \sum_{i=1}^{n} i x_i^4 + random(0,1)$ | $[-1.28,1.28]^n$ | $n=30$ | $\mathbf{x}^* = (0,\ldots,0);$ $f(\mathbf{x}^*) = 0$ |
| Dixon & Price | $f_8(\mathbf{x}) = (x_1 - 1)^2 + \sum_{i=1}^{n} i \left( 2x_i^2 - x_{i-1} \right)^2$ | $[-10,10]^n$ | $n=30$ | $\mathbf{x}^* = (0,\ldots,0);$ $f(\mathbf{x}^*) = 0$ |
| Levy | $f_9(\mathbf{x}) = 0.1\left\{\begin{array}{l}\sin^2(3\pi x_1) \\ +\sum_{i=1}^{n}(x_i-1)^2\left[1+\sin^2(3\pi x_i+1)\right] \\ +(x_n-1)^2\left[1+\sin^2(2\pi x_n)\right]\end{array}\right\}$ $+\sum_{i=1}^{n} u(x_i,5,100,4);$ $u(x_i,a,k,m) = \begin{cases} k(x_i-a)^m & x_i > a \\ 0 & -a < x_i < a \\ k(-x_i-a)^m & x_i < -a \end{cases}$ | $[-10,10]^n$ | $n=30$ | $\mathbf{x}^* = (1,\ldots,1);$ $f(\mathbf{x}^*) = 0$ |
| Sum of Squares | $f_{10}(\mathbf{x}) = \sum_{i=1}^{n} i x_i^2$ | $[-10,10]^n$ | $n=30$ | $\mathbf{x}^* = (0,\ldots,0);$ $f(\mathbf{x}^*) = 0$ |
| Zakharov | $f_{11}(\mathbf{x}) = \sum_{i=1}^{n} x_i^2 + \left(\sum_{i=1}^{n} 0.5 i x_i\right)^2 + \left(\sum_{i=1}^{n} 0.5 i x_i\right)^4$ | $[-5,10]^n$ | $n=30$ | $\mathbf{x}^* = (0,\ldots,0);$ $f(\mathbf{x}^*) = 0$ |
| Penalized | $f_{12}(\mathbf{x}) = \frac{\pi}{n}\left\{\begin{array}{l}10\sin(\pi y_1) + \\ \sum_{i=1}^{n-1}(y_i-1)^2\left[1+10\sin^2(\pi y_{i+1})\right]+(y_n-1)^2\end{array}\right\}$ $+\sum_{i=1}^{n} u(x_i,10,100,4)$ $y_i = 1 + \frac{(x_i+1)}{4}$ $u(x_i,a,k,m) = \begin{cases} k(x_i-a)^m & x_i > a \\ 0 & -a \le x_i \le a \\ k(-x_i-a)^m & x_i < a \end{cases}$ | $[-50,50]^n$ | $n=30$ | $\mathbf{x}^* = (0,\ldots,0);$ $f(\mathbf{x}^*) = 0$ |
| Penalized 2 | $f_{13}(\mathbf{x}) = 0.1\left\{\begin{array}{l}\sin^2(3\pi x_1) \\ +\sum_{i=1}^{n}\begin{array}{l}(x_i-1)^2\left[1+\sin^2(3\pi x_i+1)\right] \\ +(x_n-1)^2\left[1+\sin^2(2\pi x_n)\right]\end{array}\end{array}\right\}$ $+\sum_{i=1}^{n} u(x_i,5,100,4)$ where $u(x_i,a,k,m)$ is the same as Penalized function. | $[-50,50]^n$ | $n=30$ | $\mathbf{x}^* = (0,\ldots,0);$ $f(\mathbf{x}^*) = 0$ |
| Schwefel | $f_{14}(\mathbf{x}) = \sum_{i=1}^{n} -x_i \sin\left(\sqrt{|x_i|}\right)$ | $[-500,500]^n$ | $n=30$ | $\mathbf{x}^* = (420,\ldots,420);$ $f(\mathbf{x}^*) = -418.9829 \times n$ |
| Rastrigin | $f_{15}(\mathbf{x}) = \sum_{i=1}^{n}\left[x_i^2 - 10\cos(2\pi x_i) + 10\right]$ | $[-5.12,5.12]^n$ | $n=30$ | $\mathbf{x}^* = (0,\ldots,0);$ $f(\mathbf{x}^*) = 0$ |
| Ackley | $f_{16}(\mathbf{x}) = -20\exp\left(-0.2\sqrt{\frac{1}{n}\sum_{i=1}^{n} x_i^2}\right) -$ $\exp\left(\frac{1}{n}\sum_{i=1}^{n}\cos(2\pi x_i)\right) + 20 + \exp$ | $[-32,32]^n$ | $n=30$ | $\mathbf{x}^* = (0,\ldots,0);$ $f(\mathbf{x}^*) = 0$ |
| Griewank | $f_{17}(\mathbf{x}) = \frac{1}{4000}\sum_{i=1}^{n} x_i^2 - \prod_{i=1}^{n}\cos\left(\frac{x_i}{\sqrt{i}}\right) + 1$ | $[-600,600]^n$ | $n=30$ | $\mathbf{x}^* = (0,\ldots,0);$ $f(\mathbf{x}^*) = 0$ |
| Powelll | $f_{18}(\mathbf{x}) = \sum_{i=1}^{n/k} (x_{4i-3} + 10x_{4i-2})^2 + 5(x_{4i-1} - x_{4i})^2$ $+ (x_{4i-2} - x_{4i-1})^4 + 10(x_{4i-3} - x_{4i})^4$ | $[-4,5]^n$ | $n = 30$ | $\mathbf{x}^* = (0,\ldots,0);$ $f(\mathbf{x}^*) = 0$ |
| Salomon | $f_{19}(\mathbf{x}) = -\cos\left(2\pi\sqrt{\sum_{i=1}^n x_i^2}\right) + 0.1\sqrt{\sum_{i=1}^n x_i^2} + 1$ | $[-100,100]^n$ | $n = 30$ | $\mathbf{x}^* = (0,\ldots,0);$ $f(\mathbf{x}^*) = 0$ |

**Table A.** Test functions used in the experimental study.

Please cite this article as:
**Cuevas, E., Cienfuegos, M., Zaldívar, D., Pérez-Cisneros, M. A swarm optimization algorithm inspired in the behavior of the social-spider,** *Expert Systems with Applications*, 40 (16), (2013), pp. 6374-6384[16] Uetz, G. W. Colonial web-building spiders: Balancing the costs and. In E. J. Choe and B. Crespi, The Evolution of Social Behavior in Insects and Arachnids (pp. 458–475). Cambridge, England.: Cambridge University Press.

[17] Aviles, L. Sex-Ratio Bias and Possible Group Selection in the Social Spider Anelosimus eximius. The American Naturalist, 128(1), (1986), 1-12.

[18] Burgess, J. W. Social spacing strategies in spiders. In P. N. Rovner, Spider Communication: Mechanisms and Ecological Significance (pp. 317–351). Princeton, New Jersey.: Princeton University Press, (1982).

[19] Maxence, S. Social organization of the colonial spider Leucauge sp. in the Neotropics: vertical stratification within colonies. The Journal of Arachnology 38, (2010), 446–451.

[20] Eric C. Yip, K. S. Cooperative capture of large prey solves scaling challenge faced by spider societies. Proceedings of the National Academy of Sciences of the United States of America, 105(33), (2008), 11818-11822.

[21] Oster, G., Wilson, E. Caste and ecology in the social insects. Princeton, N.J. Princeton University press, 1978.

[22] Bert Hölldobler, E.O. Wilson. Journey to the Ants: A Story of Scientific Exploration, 1994, ISBN 0-674-48525-4.

[23] Bert Hölldobler, E.O. Wilson: "The Ants, Harvard University Press, 1990, ISBN 0-674-04075-9.

[24] Avilés, L. Causes and consequences of cooperation and permanent-sociality in spiders. In B. C. Choe., The Evolution of Social Behavior in Insects and Arachnids (pp. 476–498). Cambridge, Massachusetts.: Cambridge University Press, 1997.

[25] Rayor, E. C. Do social spiders cooperate in predator defense and foraging without a web? Behavioral Ecology & Sociobiology, 65(10), 2011, 1935-1945.

[26] Gove, R., Hayworth, M., Chhetri, M., Rueppell, O. Division of labour and social insect colony performance in relation to task and mating number under two alternative response threshold models, Insect. Soc. 56(3), (2009), 19–331.

[27] Ann L. Rypstra, R. S. Prey Size, Prey Perishability and Group Foraging in a Social Spider. Oecologia 86, (1), (1991), 25-30.

[28] Pasquet, A. Cooperation and prey capture efficiency in a social spider, Anelosimus eximius (Araneae, Theridiidae). Ethology 90, (1991), 121–133.

[29] Ulbrich, K., Henschel, J. Intraspecific competition in a social spider, Ecological Modelling, 115(2–3), (1999), 243-251.

[30] Jones, T., Riechert, S. Patterns of reproductive success associated with social structure and microclimate in a spider system, Animal Behaviour, 76(6), (2008), 2011-2019.

[31] Damian O., Andrade, M., Kasumovic, M. Dynamic Population Structure and the Evolution of Spider Mating Systems, Advances in Insect Physiology, 41, (2011), 65-114.

[32] Yang X-S (2008) Nature-inspired metaheuristic algorithms. Luniver Press, Beckington.20

# MATLAB SOFTWARE

**Case 1**
The software contains a main function SSO.m and five auxiliary functions (FeMove.m, Griewank.m, MaMove.m, Mating.m and Survive.m). Copy all files ina subdirectory and run SSO.m.

SSO.m implements an optimization example (Griewank function) which can be modified.

**Case 2**

Decompress the file SSO.rar and run SSO.m.